\documentclass{article}
\usepackage{graphicx} 
\usepackage{xcolor} 
\usepackage{makecell}
\usepackage{multirow}
\usepackage{amsmath, amssymb, amsthm}
\usepackage{algorithm}
\usepackage{algorithmic}
\usepackage{bbm} 
\usepackage{float}
\usepackage[numbers]{natbib}
\usepackage{appendix}
\usepackage{afterpage}

\usepackage{microtype}
\usepackage{subcaption}
\usepackage{booktabs} 

\usepackage{hyperref}
\usepackage{xurl} 

\usepackage[accepted]{icml2026}
\usepackage{mathtools}

\usepackage[capitalize,noabbrev]{cleveref}

\theoremstyle{plain}

\theoremstyle{definition}

\theoremstyle{remark}

\newcommand{\Reflect}{\operatorname{Reflect}}

\usepackage[disable,textsize=tiny]{todonotes}

\icmltitlerunning{Learning Permutation Distributions via Reflected Diffusion on Ranks}

\newcommand{\methodname}{Soft-Rank Diffusion}
\definecolor{newcolor}{RGB}{0,82,204}  

\newcommand{\SD}{SymmetricDiffusers}
\newcommand\KT{Kendall\text{-}Tau}

\DeclareMathOperator{\LiftToGrid}{LiftToGrid}
\newcommand{\argsort}{\operatorname{argsort}}

\begin{document}

\twocolumn[
  \icmltitle{Learning Permutation Distributions via Reflected Diffusion on Ranks}



  \icmlsetsymbol{equal}{*}

  \begin{icmlauthorlist}
    \icmlauthor{Sizhuang He}{equal,cs}
    \icmlauthor{Yangtian Zhang}{equal,cs}
    \icmlauthor{Shiyang Zhang}{cs}
    \icmlauthor{David van Dijk}{cs}
  \end{icmlauthorlist}

  \icmlaffiliation{cs}{Department of Computer Science, Yale University, New Haven, CT, USA}

  \icmlcorrespondingauthor{David van Dijk}{david.vandijk@yale.edu}
  \icmlcorrespondingauthor{Sizhuang He}{sizhuang.he@yale.edu}

  \icmlkeywords{Machine Learning, ICML}

  \vskip 0.3in
]



\printAffiliationsAndNotice{\icmlEqualContribution}

\begin{abstract}
    The finite symmetric group $S_N$ provides a natural domain for permutations, yet learning probability distributions on $S_N$ is challenging due to its factorially growing size and discrete, non-Euclidean structure. Recent permutation diffusion methods define forward noising via shuffle-based random walks (e.g., riffle shuffles) and learn reverse transitions with Plackett–Luce (PL) variants, but the resulting trajectories can be abrupt and increasingly hard to denoise as $N$ grows. We propose \emph{\methodname{}}, a discrete diffusion framework that replaces shuffle-based corruption with a structured soft-rank forward process: we lift permutations to a continuous latent representation of order by relaxing discrete ranks into soft ranks, yielding smoother and more tractable trajectories. For the reverse process, we introduce \emph{contextualized generalized Plackett--Luce (cGPL)} denoisers that generalize prior PL-style parameterizations and improve expressivity for sequential decision structures. Experiments on sorting and combinatorial optimization benchmarks show that Soft-Rank Diffusion consistently outperforms prior diffusion baselines, with particularly strong gains in long-sequence and intrinsically sequential settings.
\end{abstract}

\section{Introduction}

Permutations are fundamental objects in a wide range of applications, from ranking search results and recommendations \cite{feng2021revisitrecommenderpermutationprospective, matrixforrecommemder} to sorting and learning-to-rank \cite{liu2009ltr} and combinatorial optimization problems such as the traveling salesperson problem (TSP). In many settings, we aim to model a \emph{distribution} over permutations---and ultimately build generative models with which to sample permutations. 

In continuous Euclidean domains, diffusion models \cite{sohldickstein2015deepunsupervisedlearningusing, ho2020denoisingdiffusionprobabilisticmodels, song2021scorebasedgenerativemodelingstochastic,song2022denoisingdiffusionimplicitmodels} have emerged as powerful and scalable tools for modeling complex distributions like images, and related ideas have been extended beyond $\mathbb{R}^d$ to discrete objects such as text and graphs, a class of models known as Discrete Diffusion models \cite{austin2023structureddenoisingdiffusionmodels, lou2024discretediffusionmodelingestimating, dieleman2022continuousdiffusioncategoricaldata,gat2024discreteflowmatching,
gulrajani2023likelihoodbaseddiffusionlanguagemodels, vignac2023digressdiscretedenoisingdiffusion}. Yet permutations remain a particularly challenging case: although they are discrete, the state space grows factorially with sequence length $N$, and the natural transitions on permutations, like card-shuffling, are often \emph{abrupt}---small local moves can cause discontinuous, non-differentiable changes in the induced ordering. Consequently, diffusion-style constructions that work well for other discrete domains can become brittle on permutations and scale poorly with $N$.

Developing discrete diffusion models over permutations remains relatively underexplored. Recent progress on permutation diffusion~\cite{zhang2025symmetricdiffuserslearningdiscretediffusion} tackles above challenge by defining the forward noising process directly on permutations---typically as a random walk induced by \emph{riffle shuffles} \cite{gilbert1955theory}---and parameterizing the reverse dynamics with the Plackett--Luce distribution (PL) \cite{plackett,luce} and its generalizations (GPL) \cite{zhang2025symmetricdiffuserslearningdiscretediffusion}. While effective on small instances, these discrete forward trajectories can be highly ``jumpy'': even a single riffle--shuffle can simultaneously move many items across the sequence, yielding abrupt, non-smooth changes in the ordering. Performance often degrades rapidly and can even collapse in the long-$N$ regime.

In this paper, we take a different perspective: rather than diffusing \emph{within} $S_N$, we lift a permutation---viewed as the rank of each element---to a continuous latent representation and diffuse in that space. Concretely, we represent an ordering by assigning each item a continuous-valued \emph{soft rank} $z\in[0,1]$, obtaining a soft rank vector $Z\in[0,1]^N$ and recover the induced permutation by sorting, $\sigma=\argsort(Z)$. This relaxation lets us define smooth stochastic dynamics in a continuous space, while still producing a discrete permutation at any time via a simple sorting operation.

Building on this idea, we introduce \textbf{\methodname{}}, which defines the forward noising process as a reflected diffusion bridge~\cite{lou2023reflecteddiffusionmodels,xie2024reflectedflowmatching} on $[0,1]^N$, and observes permutations through the induced ordering $\sigma_t=\argsort(Z_t)$. The latent construction also yields a reverse-time sampler. We augment the \emph{intractable} discrete reverse step with an auxiliary, \emph{tractable} continuous update in the soft-rank space, followed by a projection back to permutations via sorting. We further introduce \textbf{contextualized GPL (cGPL)} and a \textbf{pointer-cGPL} variant, \emph{generalizations} of PL/GPL that make stagewise logits depend on the evolving prefix and the shrinking candidate set---a natural fit for intrinsically sequential and dynamic tasks such as TSP.

We evaluate \methodname{} on standard permutation generation benchmarks, including 4-digit MNIST sorting, TSP, and CIFAR-10~\citep{krizhevsky2009learning} jigsaw puzzles. Across settings, our method consistently outperforms prior permutation diffusion baselines and differentiable sorting baselines, with gains that widen as sequence lengths increase. Taken together, our results suggest that reflected diffusion in soft-rank space provides a principled and scalable route to permutation generative modeling.

\paragraph{Contributions.}
Our main contributions are:
\begin{itemize}
    \item We introduce \textbf{\methodname{}}, a permutation diffusion framework induced by reflected diffusion bridges in a relaxed continuous soft-rank space.
    \item We derive a \textbf{hybrid reverse sampler} that augments the intractable reverse dynamics in permutation space with tractable updates in the soft-rank space.
    \item We propose \textbf{cGPL} and \textbf{pointer-cGPL}, generalizations of PL/GPL that improve prefix-conditional expressivity for sequential permutation tasks.
    \item We demonstrate strong empirical performance on long-sequence MNIST sorting and TSP benchmarks, particularly in the large-$N$ regime.
\end{itemize}

\section{Related Works}
\paragraph{Discrete Diffusion Models.}
Diffusion models \citep{sohldickstein2015deepunsupervisedlearningusing, ho2020denoisingdiffusionprobabilisticmodels,song2021scorebasedgenerativemodelingstochastic} were initially developed for continuous data. D3PM \cite{austin2023structureddenoisingdiffusionmodels} extended this framework to discrete domains by defining forward noising processes based on masking or uniform replacement. Subsequent work introduced alternative parameterizations and training objectives for discrete diffusion \cite{lou2024discretediffusionmodelingestimating, gat2024discreteflowmatching} and further extended discrete (both in state space and time) diffusion to continuous-time variants \cite{campbell2022continuoustimeframeworkdiscrete, sun2023scorebasedcontinuoustimediscretediffusion, shi2025simplifiedgeneralizedmaskeddiffusion}. However, most existing methods assumes a manageable or factorized state space; directly extending them to permutation-valued data is challenging because the group $S_N$ has size $N!$, making naive transition representations intractable without exploiting additional structure.

\paragraph{Learning Permutations.}
Sorting algorithms can be viewed as producing a permutation of a set of items and are well understood; however, their hard discrete decisions are \emph{non-differentiable}, which hinders end-to-end training when permutations are produced by, or embedded within, neural models.
This motivates differentiable sorting methods that replace discrete swaps or permutation operators with smooth surrogates, yielding a \emph{soft} permutation (or rank) representation amenable to gradient-based optimization \cite{mena2018learninglatentpermutationsgumbelsinkhorn, cuturi2019differentiablerankssortingusing, prillo2020softsortcontinuousrelaxationargsort, blondel2020fastdifferentiablesortingranking, grover2019neuralsort, petersen2022monotonicdifferentiablesortingnetworks}.

Another widely used family is \textbf{Pointer Networks} (Ptr-Nets) \citep{vinyals2017pointernetworks}, which parameterize permutation-valued outputs by decoding a sequence of indices into the input, rather than symbols from a fixed vocabulary.
At each decoding step, an attention mechanism induces a categorical distribution over input positions, and the chosen index specifies which input element is appended next in the output ordering.
Because the effective output domain scales with input length, Ptr-Nets naturally accommodate variable-sized instances and provide a natural generic parameterization for combinatorial prediction problems whose outputs are orderings (e.g., sorting and routing).

\paragraph{Discrete Diffusion on Permutations.}
\citet{zhang2025symmetricdiffuserslearningdiscretediffusion} introduce \emph{SymmetricDiffusers}, which formulates discrete diffusion directly on the finite symmetric group $S_N$.
The forward noising dynamics are instantiated as a random walk on $S_N$, with the \emph{riffle shuffle} \cite{gilbert1955theory} serving as an effective transition that mixes rapidly to enable short diffusion chains in practice.

For the reverse (denoising) parameterization, Plackett--Luce-style (PL) models are adopted and extended to a generalized Plackett--Luce (GPL) family. While the PL distribution is defined as
\begin{equation}
    p_{\mathrm{PL}}(\sigma)=\prod_{i=1}^{N}
\frac{\exp\!\big(s_{\sigma(i)}\big)}
{\sum_{j=i}^{N}\exp\!\big(s_{\sigma(j)}\big)},
\end{equation}

i.e., it assigns a single preference score $s_k$ to each item $k$ and samples items \emph{without replacement} according to these fixed scores, the GPL family generalizes this construction by assigning step-specific scores for each item rather than a single fixed global score. As a result, GPL is strictly more expressive than PL; in particular, GPL is universal in the sense that it can represent arbitrary distributions over $S_N$.

\paragraph{Reflected Diffusion and Reflected Flow Matching.}
Diffusion models are often trained on data supported on bounded domains (e.g., $[0,255]$ for unnormalized image pixels), yet the learned dynamics can leave the domain at intermediate time steps---a pathology that is commonly handled by ad-hoc clipping. \citet{lou2023reflecteddiffusionmodels} address this issue by formulating generation as a \emph{reflected} score-based SDE, augmenting the dynamics with an additional reflection term that enforces the state constraint and keeps trajectories in-bounds throughout sampling. \citet{xie2024reflectedflowmatching} later extend the same principle to flow matching, incorporating reflection into the corresponding flow/ODE formulations. 

\section{Methods}
Let $[N]\coloneqq\{1,\dots,N\}$. A permutation is a bijection $\sigma:[N]\to[N]$.
We can write it as
\[
\sigma \;=\;
\begin{pmatrix}
1 & 2 & \cdots & N \\
\sigma(1) & \sigma(2) & \cdots & \sigma(N)
\end{pmatrix},
\]
meaning that each element $i\in[N]$ is mapped to $\sigma(i)$.
We can make the source indices $(1,2,\dots,N)$ implicit and use the one-line notation
\[
\sigma \;=\; (\sigma(1),\,\sigma(2),\,\dots,\,\sigma(N)).
\]
Under this convention, $\sigma(i)$ can be interpreted as the (destination) position of element $i$ after reordering, i.e., the \textbf{rank} of element $i$ of the original sequence in the permuted ordering.

We denote by $\mathcal{S}_N$ the set of all such permutations (i.e., the symmetric group under composition).
Applying a permutation to an ordered list simply reindexes its entries.
Equivalently, each $\sigma\in\mathcal{S}_N$ can be represented by a permutation matrix $P_\sigma\in\{0,1\}^{N\times N}$, so that the same reindexing can be implemented via matrix multiplication.

Given an instance $X=(x_1,\dots,x_N)\in\mathbb{R}^{N\times d}$ consisting of $N$ items (with associated features), we aim to model a conditional distribution over permutations $\sigma\in\mathcal{S}_N$.
We observe i.i.d.\ training pairs $(X,\sigma_0)\sim p^\star$, and our goal is to learn a generative model that can efficiently sample valid permutations conditioned on $X$. We write $X_t=\sigma_t(X)$ to be the permuted instance at time $t$.

We adopt a diffusion-style approach on $\mathcal{S}_N$.
Specifically, we define a forward noising process on permutations that gradually transforms $\sigma_0$ toward a target distribution, and train a parameterized reverse-time model $p_\theta(\sigma_{t-\Delta t}\mid\sigma_t, X)$ to approximately invert this corruption process.
At test time, sampling starts from the reference distribution and iteratively applies the learned reverse dynamics to generate $\hat{\sigma}_0\in\mathcal{S}_N$ conditioned on $X$.

\subsection{Forward Process: \methodname{}}

Instead of defining diffusion dynamics directly on the discrete space $\mathcal{S}_N$ (e.g., via card shuffling methods as in \citet{zhang2025symmetricdiffuserslearningdiscretediffusion}, which can be abrupt and unstructured), we adopt a continuous latent viewpoint in which a permutation is encoded by the relative ordering of real-valued coordinates.

The key idea is to relax the discrete rank values in $[N]$ to continuous \emph{soft ranks} in $[0,1]$, and to define a diffusion process over these continuous variables.
At any intermediate time, the induced ordering of the latent coordinates yields a valid permutation of the original items.

This relaxation brings two immediate benefits. First, it provides a Euclidean state space where noise injection and interpolation are natural, enabling diffusion-style modeling without relying on discrete, jump-like transitions.
Second, the forward marginals can be designed to be analytically tractable, which later allows us to derive a reverse-time sampler.

Intuitively, one may view each item as a particle undergoing Brownian motion: starting from an initial ordered configuration, the particles gradually diffuse, and the permutation at time $t$ is given by the snapshot ordering of their positions.

Formally, to initialize a continuous latent state, we convert a permutation $\sigma\in\mathcal{S}_N$ into a canonical \emph{soft-rank} vector in $[0,1]^N$ by mapping discrete ranks to a fixed grid.
Let $\{g_r\}_{r=1}^N$ be a uniform grid on $[0,1]$, e.g.,
\begin{equation}
g_r \;=\; \frac{r-1}{N-1}\qquad (r=1,\dots,N).
\label{eq:canonical_grid}
\end{equation}
We define the grid-mapping operator $\LiftToGrid:\mathcal{S}_N\to[0,1]^N$ by
\begin{equation}
\label{eq:map_to_grid_def}
\LiftToGrid(\sigma)_i \;\coloneqq\; g_{\sigma(i)},\qquad i\in[N].
\end{equation}
In particular, we set $Z_0\coloneqq \LiftToGrid(\sigma_0)$.

We generate $Z_t$ by coupling the data latent $Z_0$ to a randomly sampled endpoint $Z_1\sim p_{\mathrm{ref}}$ via a Brownian bridge, where $p_{\mathrm{ref}}$ is a tractable, data-independent reference distribution (e.g., $\mathcal{N}(0,I_N)$ or $\mathrm{Unif}([0,1]^N)$).
We apply this construction \emph{coordinate-wise}: each entry of $Z_t$ evolves as an independent one-dimensional bridge driven by its own Brownian motion.
For clarity, we describe a single coordinate using lowercase notation. Specifically, we consider the following SDE, which corresponds to a VE diffusion bridge as in \citet{zhou2023denoisingdiffusionbridgemodels}:
\begin{equation}
    dz_t \;=\; \frac{z_1 - z_t}{1-t}\,dt \;+\; \eta dw_t \;+\; dl_t,
    \label{eq:forward_bridge}
\end{equation}
where $w_t$ is a standard one-dimensional Brownian motion, $\eta$ is the noise scale and $l_t$ is a reflection term \cite{lou2023reflecteddiffusionmodels, xie2024reflectedflowmatching} that enforces the state constraint $z_t\in[0,1]$.

Then, at any time $t$, we obtain a discrete permutation in the rank representation simply by ranking the coordinates of $Z_t$.
Assuming distinct coordinates (almost surely under continuous noise) and using $1$-based indexing, we recover $\sigma_t$ as
\[
\sigma_t \;=\; \operatorname{argsort}\!\big(\operatorname{argsort}(Z_t)\big).
\]

\begin{figure*}[t]
  \centering
  \includegraphics[width=\textwidth]{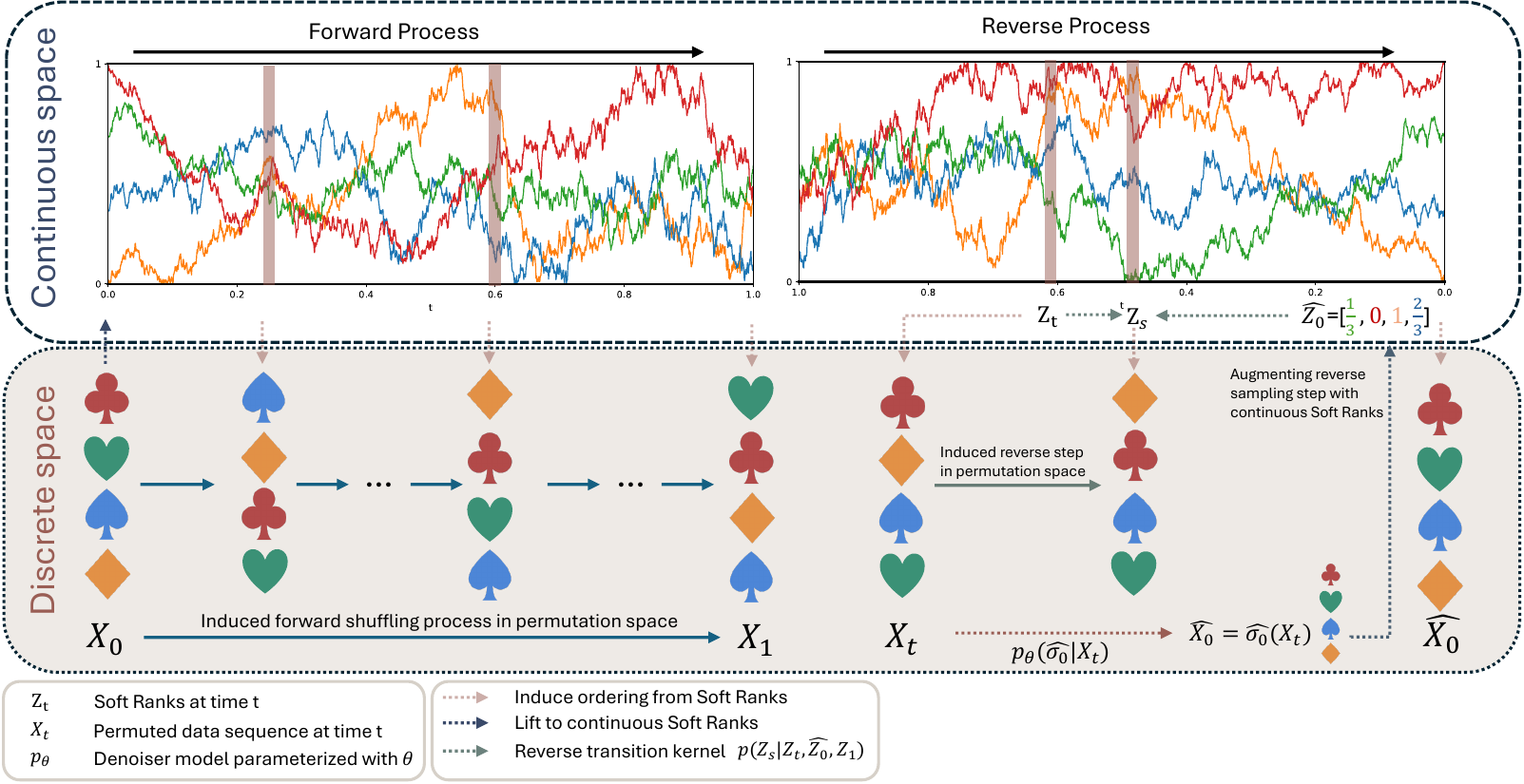}
  \caption{\textbf{\methodname{}.} We define the forward diffusion by relaxing each item's rank to a continuous \emph{soft-rank} variable and evolving these soft ranks in a reflected diffusion process \cite{lou2023reflecteddiffusionmodels}. At each time $t$, the soft ranks induce a discrete ordering by simple sorting, thereby yielding a forward process in permutation space. For reverse sampling, we couple a discrete denoiser in permutation space (predicting a clean permutation from $X_t$) with an auxiliary continuous update: we lift the predicted permutation to a grid-aligned soft-rank vector $\hat Z_0$, sample an intermediate latent $Z_s$ for $s<t$ from a conditional reverse kernel $p(Z_s \mid Z_t, \hat Z_0, Z_1)$, and map back to permutation space by sorting $Z_s$, thus stepping backward in time.}

\label{fig:soft_rank_diffusion_overview}
\end{figure*}

Figure~\ref{fig:soft_rank_diffusion_overview} illustrates the forward process.

\subsection{Reverse Process}
\label{sec:reverse_process}

Although the forward process is defined in a continuous latent space, our observations and learning targets are the induced permutations. In other words, the model operates on discrete permutation-valued states, while the continuous latent dynamics serve as an underlying construction that we invoke only to derive a convenient reverse-time update. One can choose to treat the continuous latent $Z_t$ as the primary reverse-time state and learn a denoiser directly in $\mathbb{R}^N$. We instead work in $\mathcal{S}_N$: during sampling we keep the permutation $\sigma_t$ as the explicit state, and reconstruct a rank-consistent latent $Z_t$ only as an auxiliary variable when executing each backward update.

We adopt a $\sigma_0$-prediction style parameterization. Specifically, a neural network $f_\theta$ predicts an estimate of the initial permutation,
\begin{equation}
    \hat{\sigma_0} = f_\theta(X_t,t),
    \label{eq:sample_from_model}
\end{equation}
which in turn induces an estimate of the clean sequence via the permutation action, $\hat{X}_0=\hat{\sigma}_0(X)$.

We describe the overall reverse process next, and later provide details on the parameterization of $f_\theta$.

\paragraph{Reverse-time transition kernel.}
Our forward process in Eq.~\ref{eq:forward_bridge} is a \emph{reflected} Brownian bridge. Following \citet{xie2024reflectedflowmatching}, in practice one first solve an \emph{unconstrained} backward step in $\mathbb{R}^N$ using the corresponding \emph{unreflected} bridge conditional, and apply the reflection operator only afterwards (and only when needed).

Specifically, let $\mu(u)\coloneqq (1-u)z_0+u z_1$ denote the linear interpolation between endpoints. For the \emph{unconstrained} bridge, for any $0<s<t<1$ we have the closed-form Gaussian conditional
\begin{equation}
z_s \mid z_t, z_0, z_1 \sim
\mathcal{N}\!\left(
\mu(s)+\frac{s}{t}\bigl(z_t-\mu(t)\bigr),\;
\eta^2\frac{s(t-s)}{t}
\right).
\label{eq:bridge_backward_kernel}
\end{equation}
Equivalently, we sample a proposal $\tilde z_s$ by
\begin{equation}
\tilde z_s
=\mu(s)+\frac{s}{t}\bigl(z_t-\mu(t)\bigr)
+\eta\sqrt{\frac{s(t-s)}{t}}\,\epsilon,
\qquad \epsilon\sim\mathcal{N}(0,1).
\end{equation}

and then enforce the latent-domain constraint via reflection:
\begin{equation}
z_s \leftarrow \mathcal{R}(\tilde z_s).
\end{equation}

We defer the derivation of Eq.~\eqref{eq:bridge_backward_kernel} to Appendix~\ref{sec:apdx_bridge_kernel_derivation}. 

The reverse sampling procedure is initialized by drawing $z_1\sim p_{\text{ref}}$. Since our model maintains only a permutation-valued state, we obtain the corresponding initial permutation by mapping the latent to the grid via $\sigma_1=\LiftToGrid(z_1)$ in Eq.~\eqref{eq:map_to_grid_def}. The complete reverse-time algorithm is summarized in Algorithm~\ref{alg:reverse_reflected_bridge} and illustrated in Figure~\ref{fig:soft_rank_diffusion_overview}.
\begin{algorithm}[t]
\caption{Reverse Sampling via Reflected Gaussian-Bridge Updates}
\label{alg:reverse_reflected_bridge}
\begin{algorithmic}[1]
\REQUIRE \parbox[t]{0.93\linewidth}{%
Number of items $n$; time grid $0=t_0<t_1<\cdots<t_K=1$; diffusion scale $\eta$.\\
Reference distribution $p_{\mathrm{ref}}$ on $[0,1]^n$.\\
Score network $f_\theta$ defining $p_\theta(\sigma\mid X_t,t)$.\\
Operators $\LiftToGrid(\cdot)$ and $\Reflect(\cdot)$.}

\STATE Sample $z_1 \sim p_{\mathrm{ref}}$ and set $z_{t_K}\leftarrow z_1$.
\FOR{$k=K,K-1,\dots,1$}
    \STATE $t \leftarrow t_k,\quad s \leftarrow t_{k-1}$.
    \STATE $X_t \leftarrow \argsort(z_t)$ \hfill (induce discrete state)
    \STATE Sample $\hat\sigma_0 \sim p_\theta(\cdot\mid X_t,t)$ \hfill (via $f_\theta$)
    \STATE $\hat z_0 \leftarrow \LiftToGrid(\hat\sigma_0)\in[0,1]^n$.
    \STATE $\mu_t \leftarrow (1-t)\hat z_0 + t z_1,\quad
           \mu_s \leftarrow (1-s)\hat z_0 + s z_1$.
    \STATE Sample $\xi \sim \mathcal{N}(0,I_n)$.
    \STATE $\bar z_s \leftarrow \mu_s + \frac{s}{t}\bigl(z_t-\mu_t\bigr)
           + \eta\sqrt{\frac{s(t-s)}{t}}\;\xi$.
    \STATE $z_s \leftarrow \Reflect(\bar z_s)$.
    \STATE $z_t \leftarrow z_s$.
\ENDFOR
\STATE \textbf{return} $z_{t_0}$ and optionally $\hat\sigma \leftarrow \argsort(z_{t_0})$.
\end{algorithmic}
\end{algorithm}

\subsection{Model architecture}
\label{sec:model}
We now describe the model architecture used to parameterize the $\sigma_0$-prediction distribution and to sample an estimate of the clean permutation, as in Eq.~\ref{eq:sample_from_model}.

We extend the \emph{Generalized Plackett--Luce} (GPL) parameterization of \cite{zhang2025symmetricdiffuserslearningdiscretediffusion} to a \emph{contextualized Generalized Plackett--Luce} (cGPL) model. In contrast to the encoder-only architecture used in \citet{zhang2025symmetricdiffuserslearningdiscretediffusion}, we implement cGPL with a full encoder--decoder Transformer \cite{vaswani2023attentionneed}. This formulation generalizes GPL, and we show empirically that cGPL yields a more expressive parameterization across our benchmarks.

\paragraph{Contextualized GPL.}
Similar to GPL \cite{zhang2025symmetricdiffuserslearningdiscretediffusion}, cGPL models a permutation $\sigma\in\mathcal{S}_N$ with a stagewise distribution that, at each position $i$, normalizes scores over the remaining (unselected) set
$\mathcal{R}_i(\sigma):=[N]\setminus\{\sigma(1),\ldots,\sigma(i-1)\}$:
\begin{align}
p_\theta(\sigma\mid X,t)
&=\prod_{i=1}^N \frac{\exp\!\big(s_{\sigma(i),i}\big)}{\sum_{k\in\mathcal{R}_i(\sigma)}\exp\!\big(s_{k,i}\big)}.
\label{eq:stagewise}
\end{align}

The distinction between GPL and cGPL lies entirely in the \emph{context} each position-$i$ score vector $s_{:,i}$ is allowed to depend on. In GPL, scores are prefix-agnostic and can be computed once as a static matrix. Consequently, sampling from GPL consists of a single forward pass to obtain all scores, followed by sequentially reading the corresponding columns and sampling without replacement.

\textbf{GPL:}
\begin{equation}
s_{k,i} \;\coloneqq\; [f_\theta(X,t)]_{k,i},
\label{eq:gpl_context}
\end{equation}
In contrast, cGPL makes scores explicitly prefix-conditional through an encoder--decoder Transformer. As a result, sampling from cGPL must proceed autoregressively: scores are recomputed progressively as the prefix $\sigma_{<i}$ is instantiated.

\textbf{cGPL:}
\begin{equation}
s_{k,i}=[f_\theta(X,\sigma_{<i},t)]_{k,i}.
\label{eq:cgpl_context}
\end{equation}

See Algorithm~\ref{alg:cgpl-sampling} for the complete sampling procedure of cGPL. Empirically, prefix-conditioning improves performance across all tasks we consider, and the benefits are particularly pronounced on intrinsically sequential and dynamic settings such as the Traveling Salesperson Problem, where next-step decisions depend strongly on the evolving partial solution.

When the decoder is made prefix-agnostic such that $s_{k,i}(X,\sigma_{<i},t)\equiv s_{k,i}(X,t)$ for all prefixes, cGPL reduces to GPL; hence we can conclude cGPL generalizes GPL.

\begin{figure}[t]
  \centering
  \begin{subfigure}[c]{0.44\linewidth}
    \centering
    \vspace{0pt}
    \includegraphics[width=\linewidth]{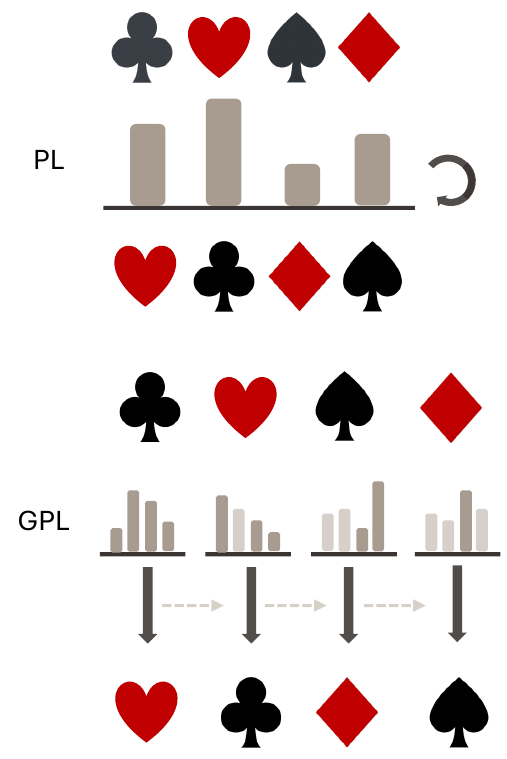}
    \caption{PL and GPL sampling.}
    \label{fig:subfig_pl_gpl_sampling}
  \end{subfigure}%
  \hfill
  \begin{subfigure}[c]{0.54\linewidth}
    \centering
    \vspace{0pt}
    \includegraphics[width=\linewidth]{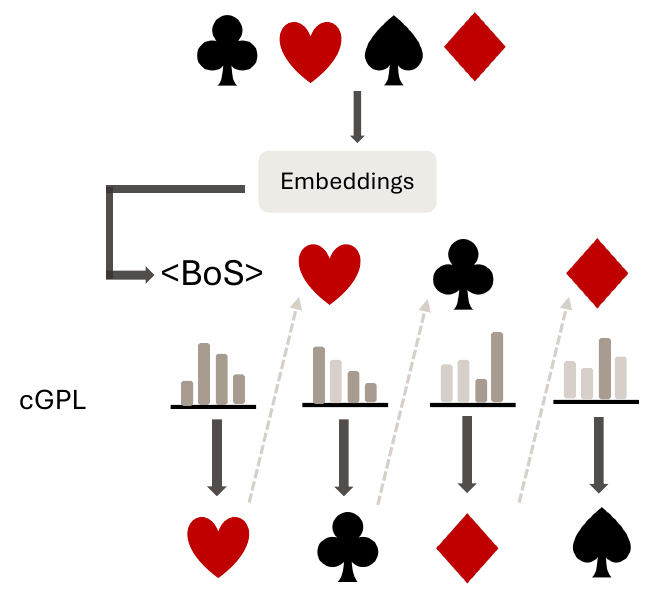}
    \caption{cGPL sampling.}
    \label{fig:subfig_cgpl_sampling}
  \end{subfigure}
  \caption{\textbf{Sampling in PL/GPL/cGPL.}
  \textbf{(a) PL and GPL.} In PL, each item is assigned a single scalar score; sampling a permutation amounts to repeatedly sampling from the same score vector without replacement, masking selected items and renormalizing at each step. In GPL, each item is assigned a length-$N$ score vector, yielding position-specific logits: at step $i$ we sample according to the $i$-th column of the score matrix (after masking previously selected items), proceeding sequentially from $i=1$ to $N$.
  \textbf{(b) cGPL.} In cGPL, each item is assigned a position-dependent score vector \emph{dynamically}. Sampling proceeds \emph{autoregressively}: the score vector at later positions depends on the outcomes sampled at preceding positions. As in GPL, we apply masking and subsequent renormalization to obtain a valid probability distribution over the remaining items.}
  \label{fig:pl_gpl_cgpl_sampling}
\end{figure}

Figure~\ref{fig:pl_gpl_cgpl_sampling} demonstrates the difference among sampling from PL, GPL and cGPL. 

In practice, both sampling and likelihood evaluation require normalizing scores only over the remaining (feasible) set $\mathcal{R}_i(\sigma)$.
We implement this via a feasibility mask $m_{k,i}(\sigma_{<i})\in\{0,-\infty\}$ added to the predicted scores:
\begin{align}
m_{k,i}(\sigma_{<i})&=
\begin{cases}
0, & k\in \mathcal{R}_i(\sigma),\\
-\infty, & k\notin \mathcal{R}_i(\sigma),
\end{cases}\\
\tilde{s}_{k,i} \;&:=\; s_{k,i}(X,\sigma_{<i},t) + m_{k,i}(\sigma_{<i}),
\label{eq:mask}
\end{align}
so that $\mathrm{softmax}(\tilde{s}_{:,i})$ renormalizes over $\mathcal{R}_i(\sigma)$ and assigns zero probability to infeasible items.

\paragraph{Training objective.}
We train our model by maximizing the standard variational lower bound:
\begin{align}
\label{eq:elbo}
&\phantom{\ge\;}
\mathbb{E}_{p_{\mathrm{data}}(X_0,X)}[\log p_\theta(X_0\mid X)]
\notag\\
\ge\;&
\mathbb{E}_{p_{\mathrm{data}}(X_0,X)\,q(X_{1:T}\mid X_0,X)}
\Big[
\log p(X_T\mid X)
\notag\\
&\hspace{2.6em}
+\sum_{t=1}^{T}\log \frac{p_\theta(X_{t-1}\mid X_t,t)}{q(X_t\mid X_{t-1})}
\Big].
\notag
\end{align}

Our model predicts a denoised distribution $p_\theta(X_0\mid X_t,t)$. To obtain a one-step reverse transition, we compose this predictor with
a reconstruction kernel $\tilde{q}_t(X_{t-1}\mid X_t,X_0)$:
\begin{equation}
p_\theta(X_{t-1}\mid X_t,t)
\ :=\
\sum_{X_0} \tilde{q}_t(X_{t-1}\mid X_t,X_0)\,p_\theta(X_0\mid X_t,t).
\label{eq:induced_reverse}
\end{equation}

Note that $\tilde{q}_t$ in the discrete permutation space is generally intractable but as discussed in \ref{sec:reverse_process}, we can augment with a tractable update step in the continuous soft-rank space.

With the feasibility mask in Eq~\ref{eq:mask}, the negative log-likelihood reduces to a sum of cross-entropies on the masked scores:
\begin{equation}
\mathcal{L}(\sigma;X,t)= -\sum_{i=1}^{N}\log\frac{\exp\!\big(\tilde{s}_{\sigma(i),i}\big)}{\sum_{k=1}^{N}\exp\!\big(\tilde{s}_{k,i}\big)}.
\label{eq:nll}
\end{equation}

\begin{algorithm}[t]
\caption{Autoregressive Sampling from cGPL}
\label{alg:cgpl-sampling}
\begin{algorithmic}[1]
\REQUIRE Input set $X=\{x_1,\ldots,x_n\}$; diffusion step $t$; neural decoder $f_\theta$
\ENSURE Permutation $\hat{\sigma}\in\mathcal{S}_n$
\STATE $\mathcal{R} \leftarrow \{1,2,\ldots,n\}$ \hfill (remaining items)
\STATE $\hat{\sigma} \leftarrow [\ ]$ \hfill (empty prefix)
\FOR{$i=1,\ldots,n$}
    \STATE Compute logits $s_{:,i} \leftarrow \big( [f_\theta(X,\hat{\sigma}_{<i},t)]_{k,i} \big)_{k=1}^n$
    \STATE $\tilde{s}_{k,i} \leftarrow s_{k,i}$ if $k\in\mathcal{R}$, else $-\infty$ \hfill (mask)
    \STATE $\pi_{\cdot,i} \leftarrow \mathrm{softmax}(\tilde{s}_{\cdot,i})$
    \STATE Sample $\hat{\sigma}(i) \sim \mathrm{Categorical}(\pi_{\cdot,i})$
    \STATE $\mathcal{R} \leftarrow \mathcal{R}\setminus\{\hat{\sigma}(i)\}$
\ENDFOR
\STATE \textbf{return} $\hat{\sigma}$
\end{algorithmic}
\end{algorithm}

\paragraph{Pointer parameterization.}
As an alternative to a fixed $N$-way linear head that directly outputs $N$ logits (treating output coordinates as a fixed ``vocabulary'' indexed by absolute positions), we draw inspiration from Pointer Networks \cite{vinyals2017pointernetworks} and parameterize the cGPL scores with an item-aligned pointer head that explicitly \emph{points} to the encoded input items.

Concretely, at decoding step $i$ we compute a compatibility score between the decoder state and each encoder item representation, and interpret the resulting length-$N$ vector as a categorical distribution over the \emph{input indices} (rather than over a fixed output dictionary); see Fig.~\ref{fig:model_arch} for an illustration.

This parameterization naturally supports variable-size output dictionaries and provides another way of producing valid permutations other than relying on versions of the Plackett-Luce distribution \cite{vinyals2017pointernetworks}.

\vspace{0.25em}
\noindent
Formally, given encoder item representations and a prefix-dependent decoder state, we compute the pointer scores as a bi-affine transformation between them:
\begin{align}
e_k \;&=\; \mathrm{Enc}_\theta(X_t,t)_k \in \mathbb{R}^d, \qquad k\in[N], \notag\\
d_i \;&=\; \mathrm{Dec}_\theta(\sigma_{<i},X_t,t)_i \in \mathbb{R}^d, \qquad i\in[N], \notag\\
s_{k,i}(X_t,\sigma_{<i},t)
\;&=\; d_i^\top W e_k \;+\; u^\top d_i \;+\; v^\top e_k \;+\; b,
\label{eq:biaffine_pointer}
\end{align}
where $W\in\mathbb{R}^{d\times d}$ and $u,v\in\mathbb{R}^{d}$ and $b\in\mathbb{R}$ are learned parameters.
The conditional distribution at step $i$ is then obtained by applying the same remaining-set normalization as in Eq.~\eqref{eq:stagewise}.

\paragraph{Speeding up inference with KV caching.}
The cGPL/Pointer-cGPL decoders are autoregressive: at each diffusion timestep $t$, while the encoder-only GPL model needs only a single forward pass to compute the scores, our cGPL/Pointer-cGPL decoder must run $N$ sequential forward passes, one per prefix position, to produce all stagewise scores. Without optimization, each of these passes redundantly recomputes attention keys and values for previously emitted prefix tokens, leading to a substantial wall-clock overhead that grows with $N$. To remove this redundancy, we adopt \emph{key/value (KV) caching}~\citep{pope2023efficiently}, a well-established technique in natural language modeling with autoregressive transformers, which we can apply straightforwardly in our cGPL and Pointer-cGPL decoders. We quantify the resulting inference speedup in Section~\ref{sec:experiments}. Note that this overhead is specific to inference: during training, teacher forcing makes the ground-truth prefix available at every position, so the $N$ stagewise score vectors are computed in parallel within a single forward pass.

\section{Experiments}
\label{sec:experiments}

\subsection{4-digit MNIST sorting}
We evaluate \methodname{} on four-digit MNIST sorting, a standard benchmark for permutation modeling in differentiable sorting and permutation diffusion \citep{grover2019neuralsort, petersen2022monotonicdifferentiablesortingnetworks, zhang2025symmetricdiffuserslearningdiscretediffusion}.
Each instance is a list of $N$ four-digit MNIST images (randomly ordered), and the task is to predict the permutation that sorts the list in ascending order by the underlying four-digit numbers they represent.
We report \emph{Kendall--Tau correlation} (a correlation coefficient between ranks) \cite{kendall}, \emph{accuracy} (fraction of exactly matched sorted lists), and \emph{correctness} (fraction of items placed in their correct positions). We rerun the baseline methods DiffSort \cite{petersen2022monotonicdifferentiablesortingnetworks}, Error-free DiffSort \cite{kim2024generalizedneuralsortingnetworks} and SymmetricDiffusers \cite{zhang2025symmetricdiffuserslearningdiscretediffusion} using their official GitHub repositories and the recommended (or default) hyperparameters. For SymmetricDiffusers, we use GPL sampling and disable beam search so that the comparison focuses on cGPL versus GPL under the same sampling setting. See Appendix~\ref{sec:apdx_MNIST_details} for details.

Table~\ref{tab:mnist-sorting} summarizes results on the 4-digit MNIST sorting benchmark across sequence lengths $N\in\{9,15,32,52,75,100,150,200\}$. Overall, \methodname{} consistently outperforms all baseline methods, with the performance gap widening as $N$ increases. In particular, the Pointer variant further outperforms the plain cGPL variant, and is the only method that maintains non-trivial performance in the long-sequence regime ($N\ge 150$). While SymmetricDiffusers performs strongly on short sequences, its accuracy and correctness degrade rapidly as $N$ grows, approaching chance-level behavior at $N>100$. By contrast, \methodname{} degrades more gracefully, indicating improved scalability and robustness to longer permutations. 

\begin{figure}[t]
  \centering
  \begin{subfigure}[t]{0.495\linewidth}
    \centering
    \includegraphics[width=\linewidth]{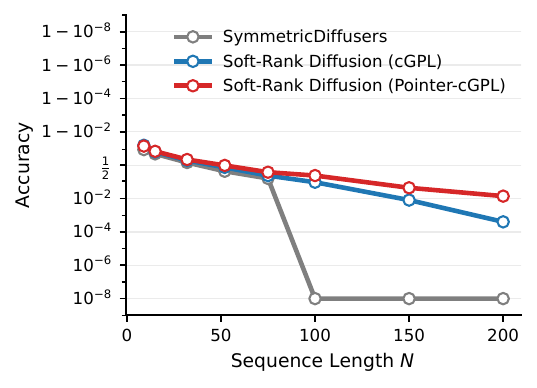}
    \caption{Exact-match accuracy in logit scale over $N$.}
    \label{fig:mnist-accuracy}
  \end{subfigure}
  \hfill
  \begin{subfigure}[t]{0.455\linewidth}
    \centering
    \includegraphics[width=\linewidth]{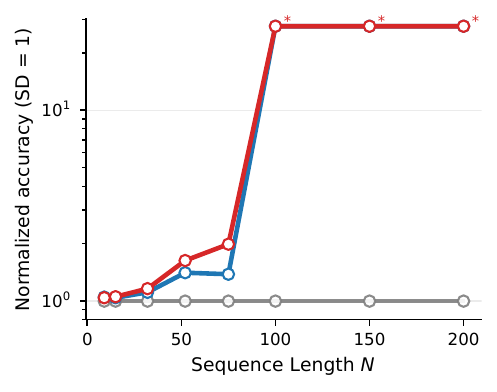}
    \caption{Accuracy ratio relative to SymmetricDiffusers (SD), shown on a log scale.}
    \label{fig:mnist-gain}
  \end{subfigure}
  \caption{\textbf{Exact-match accuracy and relative improvement versus SymmetricDiffusers as a function of sequence length (N).} Panel~\ref{fig:mnist-accuracy} shows accuracy on a logit scale. Panel~\ref{fig:mnist-gain} reports the accuracy ratio of \methodname{} relative to SymmetricDiffusers on a log scale. Starred points mark sequence lengths where SymmetricDiffusers attains zero accuracy; the corresponding ratios are undefined and are clipped to the top of the plotted range for visualization.}
  \label{fig:mnist-scaling}
\end{figure}

This scaling advantage is further illustrated in Figure~\ref{fig:mnist-scaling}, which examines exact-match accuracy as a function of the sequence length~$N$.
Figure~\ref{fig:mnist-accuracy} shows that both variants of \methodname{} are substantially less prone to catastrophic accuracy collapse as $N$ increases, with the Pointer variant exhibiting consistently stronger robustness at larger sequence lengths.
Figure~\ref{fig:mnist-gain} further quantifies this effect by reporting the relative exact-match accuracy of \methodname{} over \SD, computed as the ratio relative to the baseline accuracy.
The improvement margin grows rapidly with $N$, indicating that \methodname{} achieves increasingly favorable scaling behavior compared to \SD.

\begin{table*}[t]
\centering
\small
\caption{\textbf{MNIST sorting benchmark.} We compare baselines and our models across sequence lengths. Bold indicates the best value in each column. For a controlled comparison, \SD{} and both \methodname{} variants use the same 7-layer Transformer backbone (ours: 3-layer encoder + 4-layer decoder with matched hidden width). This differs from \citet{zhang2025symmetricdiffuserslearningdiscretediffusion}, which used a 12-layer Transformer for $n=200$ and a 7-layer Transformer for other lengths.}
\makebox[\textwidth][c]{%
\begin{tabular}{@{}p{3.5cm}lcccccccc@{}}
\toprule
Method & Metric & \multicolumn{8}{c}{Sequence Length $N$} \\
\cmidrule(lr){3-10}
& & 9 & 15 & 32 & 52 & 75 & 100 & 150 & 200 \\
\midrule
\multirow{3}{*}{\shortstack[l]{DiffSort \\ \cite{petersen2022monotonicdifferentiablesortingnetworks}}}
& \KT{} $\uparrow$      & 0.8163 & 0.7386 & 0.4423 & 0.3037 & 0.2227 & 0.1617 & 0.0891 & 0.0636 \\
& Accuracy $\uparrow$ & 0.5555 & 0.2350 & 0.0003 & 0.0000 & 0.0000 & 0.0000 & 0.0000 & 0.0000 \\
& Correctness $\uparrow$  & 0.8643 & 0.7975 & 0.5335 & 0.3897 & 0.2987 & 0.2251 & 0.1307 & 0.0953 \\
\midrule
\multirow{3}{*}{\shortstack[l]{Error-free DiffSort \\ \cite{kim2024generalizedneuralsortingnetworks}}}
& \KT{} $\uparrow$      & 0.9151 & 0.9160 & 0.8798 & 0.0042 & 0.0005 & 0.0029 & 0.0006 & 0.0002 \\
& Accuracy $\uparrow$ & 0.7953 & 0.7492 & 0.5186 & 0.0000 & 0.0000 & 0.0000 & 0.0000 & 0.0000 \\
& Correctness $\uparrow$  & 0.9365 & 0.9334 & 0.9019 & 0.0258 & 0.0132 & 0.0112 & 0.0072 & 0.0030 \\
\midrule
\multirow{3}{*}{\shortstack[l]{SymmetricDiffusers \\ 
with GPL\\
\cite{zhang2025symmetricdiffuserslearningdiscretediffusion}}} 
& \KT{} $\uparrow$      & 0.9483 & 0.9320 & 0.8610 & 0.7890 & 0.7304 & 0.1114 & 0.0641 & 0.0022 \\
& Accuracy $\uparrow$ & 0.8955 & 0.8260 & 0.5896 & 0.3009 & 0.1379 & 0.0000 & 0.0000 & 0.0000 \\
& Correctness $\uparrow$  & 0.9605 & 0.9450 & 0.8846 & 0.8232 & 0.7733 & 0.1620 & 0.0975 & 0.0061 \\
\midrule
\multirow{3}{*}{\shortstack[l]{\methodname{} \\ with cGPL \\ \textbf{Ours}}}
& \KT{} $\uparrow$          & \textbf{0.9668}   & 0.9384            & 0.8765 & 0.8092 & 0.6925 & 0.6279 & 0.4821 & 0.3982 \\
& Accuracy $\uparrow$       & \textbf{0.9397}   & 0.8601            & 0.6555 & 0.4239 & 0.1908 & 0.0863 & 0.0080 & 0.0004 \\
& Correctness $\uparrow$    & \textbf{0.9742}   & 0.9505            & 0.8967 & 0.8393 & 0.7347 & 0.6773 & 0.5468 & 0.4644 \\
\midrule
\multirow{3}{*}{\shortstack[l]{\methodname{} \\ with Pointer-cGPL \\ \textbf{Ours}}}
& \KT{} $\uparrow$          & 0.9645            & \textbf{0.9455}   & \textbf{0.8944} & \textbf{0.8459} & \textbf{0.7727} & \textbf{0.7454} & \textbf{0.6198} & \textbf{0.6048} \\
& Accuracy $\uparrow$       & 0.9329            & \textbf{0.8710}   & \textbf{0.6854} & \textbf{0.4909} & \textbf{0.2742} & \textbf{0.1922} & \textbf{0.0421} & \textbf{0.0137} \\
& Correctness $\uparrow$    & 0.9719            & \textbf{0.9559}   & \textbf{0.9119} & \textbf{0.8724} & \textbf{0.8082} & \textbf{0.7849} & \textbf{0.6704} & \textbf{0.6607} \\
\bottomrule
\end{tabular}%
}

\label{tab:mnist-sorting}
\end{table*}

\subsection{Traveling Salesperson Problem (TSP)}
We evaluate \methodname{} on the traveling salesperson problem (TSP), an NP-hard combinatorial optimization task. Given a set of 2D points $V=\{v_1,\ldots,v_N\}\subset\mathbb{R}^2$, the goal is to predict a tour permutation $\sigma\in S_N$ that minimizes the total length $\sum_{i=1}^{N}\lVert v_{\sigma(i)}-v_{\sigma(i+1)}\rVert_2$, where $\sigma(N+1):=\sigma(1)$. We report the tour length and the optimality gap (see Appendix~\ref{sec:apdx_TSP_details} for details) on TSP-20 and TSP-50 ($N=20$ and $N=50$, respectively). We rerun \SD{} using their respective official GitHub repositories and default hyper-parameters. For SymmetricDiffusers, we again use GPL sampling and disable beam search to ensure a fair comparison. See Appendix~\ref{sec:apdx_TSP_details}

This task is particularly well-suited to highlight the advantage of our \emph{dynamic} cGPL parameterization: unlike static scoring schemes, TSP decoding requires making prefix-dependent decisions over a shrinking set of remaining cities, where the desirability of each candidate depends strongly on the current partial tour. By producing context-conditioned, step-specific logits over the remaining items, cGPL better matches this dynamic decision structure and therefore exhibits larger performance gains as problem size increases.

\begin{table}[t]
\centering
\footnotesize
\setlength{\tabcolsep}{4pt}
\caption{\textbf{TSP performance on TSP-20 and TSP-50.} We report tour length $L(\text{Tour})$ and the optimality gap (both lower is better).}
\begin{tabular}{@{}p{3.0cm}cccc@{}}
\toprule
Method
& \multicolumn{2}{c}{TSP-20}
& \multicolumn{2}{c}{TSP-50} \\
\cmidrule(lr){2-3}\cmidrule(lr){4-5}
& $L(\text{Tour})$ $\downarrow$ & Gap $\downarrow$
& $L(\text{Tour})$ $\downarrow$ & Gap $\downarrow$ \\
\midrule

\shortstack[l]{SymmetricDiffusers \\ 
with GPL \\ 
\cite{zhang2025symmetricdiffuserslearningdiscretediffusion}}
& 5.4342 & 0.4169 & 12.8525 & 1.2618 \\
\midrule

\shortstack[l]{\methodname{} \\ with cGPL \\ \textbf{Ours}}
& 3.8732 & 0.0079 & 5.7336 & 0.0078 \\
\midrule

\shortstack[l]{\methodname{} \\ with Pointer-cGPL \\ \textbf{Ours}}
& \textbf{3.8557} & \textbf{0.0034} & \textbf{5.7157} & \textbf{0.0047} \\
\bottomrule
\end{tabular}

\label{tab:tsp}
\end{table}

Table~\ref{tab:tsp} reports Euclidean TSP results on TSP-20 and TSP-50. Across both problem sizes, \methodname{} substantially improves over SymmetricDiffusers, reducing tour length and closing the optimality gap by more than two orders of magnitude. Moreover, the Pointer variant consistently achieves the best performance, yielding further gains over the fixed variant—especially on TSP-50—highlighting the benefit of our dynamic cGPL parameterization for prefix-dependent decoding over a shrinking candidate set.

\subsection{CIFAR-10 Jigsaw}
\label{sec:cifar10-jigsaw}

We evaluate \methodname{} on the CIFAR-10~\citep{krizhevsky2009learning} jigsaw puzzle task as a third benchmark complementing MNIST sorting and TSP. A CIFAR-10 image is divided into an $n\times n$ grid of patches ($N=n^2$) presented in random order, and the model must predict the permutation that reconstructs the original raster ordering. Unlike sorting (element-wise) or TSP (sequential), jigsaw reconstruction requires global spatial reasoning over all patches: placing a patch depends on the full image context. We compare against \SD{}~\cite{zhang2025symmetricdiffuserslearningdiscretediffusion} at three scales, $3\times 3$, $4\times 4$, and $8\times 8$, using the same metrics as in the MNIST sorting experiments.

Table~\ref{tab:cifar10-jigsaw} reports the results. The qualitative pattern matches sorting and TSP: as $N$ grows, the gap between \methodname{} and \SD{} widens, and at $8\times 8$ the discrete baseline collapses to chance-level behavior while \methodname{} continues to learn meaningful structure. The decoder comparison at $8\times 8$ further shows the Pointer variant substantially outperforming vanilla cGPL, consistent with the intuition that cross-attention over all positions becomes critical when the permutation space is large.

\begin{table*}[t]
\centering
\small
\caption{\textbf{CIFAR-10 jigsaw benchmark.} Bold marks the best value per (metric, $n$). All methods use a 7-layer Transformer backbone with the same hidden width.}
\begin{tabular*}{\textwidth}{@{\extracolsep{\fill}}p{3.5cm}lccc@{}}
\toprule
Method & Metric & \multicolumn{3}{c}{Grid Size $n\times n$} \\
\cmidrule(lr){3-5}
& & $3\times3$ ($N\!=\!9$) & $4\times4$ ($N\!=\!16$) & $8\times8$ ($N\!=\!64$) \\
\midrule
\multirow{3}{*}{\shortstack[l]{SymmetricDiffusers \\
with GPL\\
\cite{zhang2025symmetricdiffuserslearningdiscretediffusion}}}
& \KT{} $\uparrow$        & 0.8312  & 0.6101  & 0.0250  \\
& Accuracy $\uparrow$     & 0.7002 & 0.2172 & 0.0000 \\
& Correctness $\uparrow$  & 0.8662  & 0.6753  & 0.0501  \\
\midrule
\multirow{3}{*}{\shortstack[l]{\methodname{} \\ with cGPL \\ \textbf{Ours}}}
& \KT{} $\uparrow$        & 0.9494           & 0.9320           & 0.2521           \\
& Accuracy $\uparrow$     & 0.9221          & \textbf{0.8752} & 0.0924          \\
& Correctness $\uparrow$  & 0.9564           & 0.9413           & 0.2854          \\
\midrule
\multirow{3}{*}{\shortstack[l]{\methodname{} \\ with Pointer-cGPL \\ \textbf{Ours}}}
& \KT{} $\uparrow$        & \textbf{0.9542}  & \textbf{0.9345}  & \textbf{0.4962}  \\
& Accuracy $\uparrow$     & \textbf{0.9258} & 0.8681          & \textbf{0.2700} \\
& Correctness $\uparrow$  & \textbf{0.9601}  & \textbf{0.9448}  & \textbf{0.5256}  \\
\bottomrule
\end{tabular*}
\label{tab:cifar10-jigsaw}
\end{table*}

\begin{figure}[t]
    \centering
    \includegraphics[width=\linewidth]{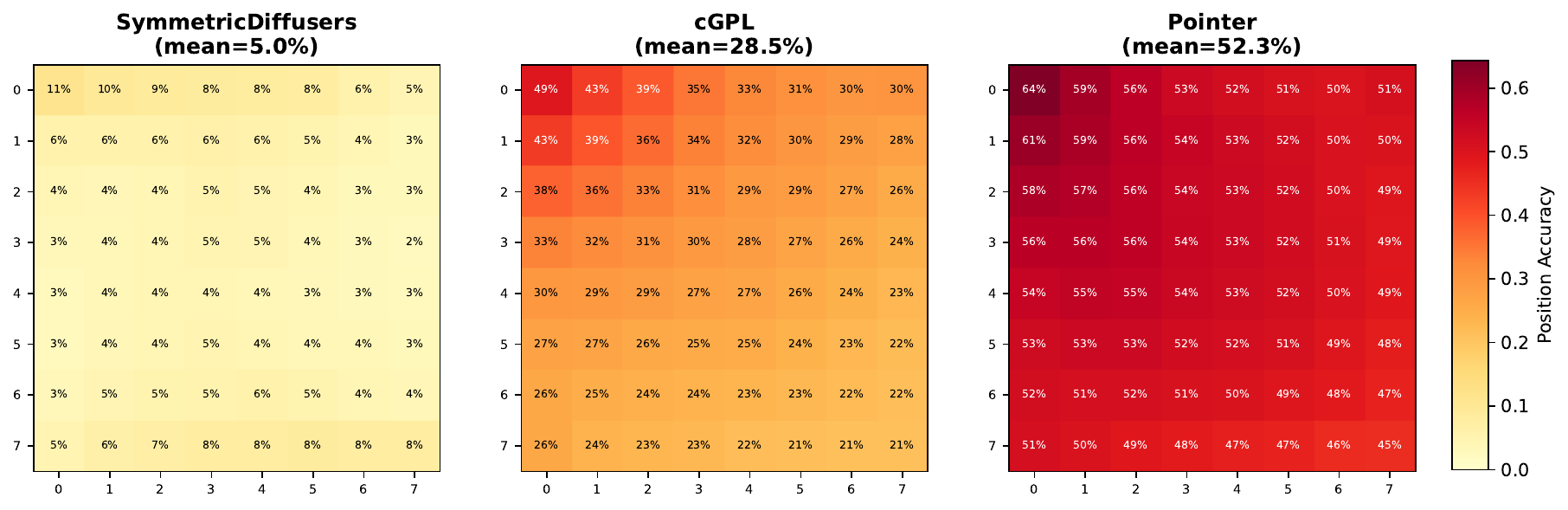}
    \caption{\textbf{Per-position accuracy heatmaps on $8\times8$ CIFAR-10 jigsaw.} Each cell is the fraction of test samples where that grid position is correctly assigned. Left: \SD{} (mean $5.0\%$). Middle: cGPL (mean $28.5\%$). Right: Pointer-cGPL (mean $52.3\%$). Random baseline $1/64\approx 1.6\%$.}       
    \label{fig:position_heatmap}
\end{figure}

\paragraph{Failure mode analysis.}
Figure~\ref{fig:position_heatmap} examines how each method fails at $8\times 8$. The discrete baseline makes predictions that are uniformly near chance across all positions, indicating that no meaningful positional structure has been learned. The cGPL variant exhibits a top-left to bottom-right accuracy gradient consistent with autoregressive error accumulation, while the Pointer variant maintains high accuracy across all positions with a much flatter gradient, owing to its biaffine cross-attention head attending to all positions simultaneously. Patch-wise displacement distributions (Fig.~\ref{fig:displacement_histogram}) reinforce this picture: \methodname{} concentrates errors near the correct position—even when wrong, patches typically land within a small displacement—whereas \SD{} produces an approximately random distribution. We attribute the contrast to the locality induced by the continuous forward process: small noise in $[0,1]^N$ swaps nearby soft ranks, so the reverse process naturally resolves local ambiguities before global ones, while discrete riffle-shuffle dynamics perform global mixing at each step and leave no local structure for the reverse process to exploit.

\paragraph{Runtime and KV cache.}
Table~\ref{tab:runtime} reports wall-clock inference time. With KV cache enabled, \methodname{} is comparable to \SD{} at small $N$ but slower at large $N$ due to the autoregressive cGPL decode; the additional runtime at $8\times 8$ nevertheless buys meaningful accuracy in a regime where \SD{} achieves none, as demonstrated in Figure~\ref{fig:accuracy_runtime_kvcache}. The KV-cache speedup grows with $N$, reflecting the per-prefix-position savings discussed in Section~\ref{sec:model}.

\section{Ablation Studies}
We study the impact of several key design choices of \methodname{} in Table~\ref{tab:ablation-table}. 
While our default configuration adopts an $\sigma_0$ parametrization for the cGPL reverse model, we observe that the model achieves consistently strong performance when using an $\sigma_{t-1}$ parametrization, as also employed in \citet{zhang2025symmetricdiffuserslearningdiscretediffusion}. 
In contrast, when paired with the Riffle Shuffle forward process, the $\sigma_0$ parametrization consistently fails, leading to a severe degradation in performance across all evaluation metrics. This may be due to the intractability of the reverse transition when the forward process is defined via a riffle shuffle. Moreover, even under the shared $\sigma_{t-1}$ parametrization, models trained with the \methodname{} forward process consistently outperform those using the Riffle Shuffle forward process. 
Finally, incorporating a biaffine pointer mechanism in the cGPL reverse model further improves performance over the vanilla cGPL variant, highlighting the benefit of explicitly modeling autoregressive selection in the reverse dynamics.

\section{Conclusions}\label{sec:conclusion}
We presented \methodname{}, a diffusion framework for permutation-valued data that bridges diffusion modeling with principled ranking distributions. By lifting permutations into a continuous latent space through soft rank representations, our approach enables a smoother and more tractable forward process than existing abrupt schemes such as riffle--shuffles, and a theoretically sound reverse solver grounded in Plackett–Luce–style likelihoods. Central to our method is the proposed contextualized Generalized Plackett–Luce (cGPL) parameterization, which conditions each selection step on the evolving prefix and generalizes prior prefix-agnostic models.

Across sorting and combinatorial optimization benchmarks, \methodname{} consistently outperforms existing permutation diffusion and differentiable sorting approaches, with particularly strong gains on long and intrinsically sequential permutations. These results highlight the importance of context-aware reverse dynamics for scalable permutation modeling. We believe this framework opens the door to further unifying diffusion-based generative modeling with structured, likelihood-based formulations over discrete combinatorial objects.

\section*{Impact Statement}\label{sec:impact}
This paper advances methods in machine learning. We do not anticipate societal impacts that require specific discussion beyond standard considerations.

\bibliographystyle{icml2026}
\bibliography{ref}

\clearpage
\appendix
\onecolumn

\section{Details on experimental setup}
\subsection{4-digit MNIST sorting}
\label{sec:apdx_MNIST_details}
\paragraph{Dataset}
Following \citet{zhang2025symmetricdiffuserslearningdiscretediffusion}, we generate the 4-digit MNIST dataset using the code from their official GitHub repository. For each experiment, we construct training sequences by sampling $60{,}000$ length-$N$ sequences of 4-digit MNIST images from \texttt{torchvision.datasets.MNIST} \cite{torchvision2016}, where $N$ is an experiment-specific parameter. We generate an additional $10{,}000$ sequences for testing. Each 4-digit image is created on the fly by first sampling four digits uniformly at random, then drawing the corresponding digit images from MNIST, and finally concatenating them into a single 4-digit image.

\paragraph{Model architecture}
For \SD{}, we use a 7-layer encoder-only Transformer with hidden dimension 512, feedforward dimension 128, with 8 attention heads. For \methodname{}, we match the overall model capacity by using a 7-layer encoder--decoder Transformer, with 3 layers in the encoder and 4 layers in the decoder. We keep the hidden and feedforward dimensions the same as \SD{}. \textbf{Note:} While \citet{zhang2025symmetricdiffuserslearningdiscretediffusion} use a 12-layer Transformer for $N{=}200$, we use a 7-layer Transformer for all $N$ to keep model capacity consistent across sequence lengths and ensure performance is comparable across settings.

\paragraph{Training}
All models are trained on a single NVIDIA H100 GPU with batch size 64 for 120 epochs. Training data are generated on the fly at each epoch. For \SD{}, we use the default hyperparameters from the authors’ official GitHub repository. For $N\in{75,150}$, where no default hyperparameters are provided, we reuse the settings for $N=52$ and $N=100$, respectively.

\paragraph{Inference and metrics}
At inference time, we randomly generate $10{,}000$ sequences of $N$ 4-digit MNIST images. Since each image is labeled by its underlying 4-digit number, we can obtain the ground-truth ascending order directly from the labels. We report the same three metrics as \citet{zhang2025symmetricdiffuserslearningdiscretediffusion}: Kendall--Tau coefficient, Accuracy (exact match), and Correctness (element-wise match). Kendall--Tau measures rank correlation between the predicted and ground-truth permutations; Accuracy indicates whether the two permutations are identical; and Correctness is the fraction of elements placed in the correct position.

\paragraph{Non-diffusion model baselines}
We retrain and evaluate DiffSort \cite{petersen2022monotonicdifferentiablesortingnetworks} and Error-free DiffSort \cite{kim2024generalizedneuralsortingnetworks} under the same dataset setup. For DiffSort, we use the default hyperparameters from the authors’ official GitHub repository, with the odd--even sorting network, steepness $10$, and learning rate $10^{-3.5}$. For Error-free DiffSort, we use TransformerL as the backbone and the hyperparameters listed in Table~\ref{tab:HP_ef_diffsort}.

\begin{table}[!htb]
  \centering
  \caption{Hyperparameters for Error-Free Diffsort on 4-digit MNIST sorting.}
  \label{tab:HP_ef_diffsort}
  \begin{tabular}{ccccc}
    \toprule
    \textbf{Sequence Length} & \textbf{Steepness} & \textbf{Sorting Network} & \textbf{Loss Weight} & \textbf{Learning Rate} \\
    \midrule
    9   & 34  & odd even & 1.00 & $10^{-4}$ \\
    15  & 25  & odd even & 0.10 & $10^{-4}$ \\
    32  & 124 & odd even & 0.10 & $10^{-4}$ \\
    52  & 130 & bitonic  & 0.10 & $10^{-3.5}$ \\
    75  & 135 & bitonic  & 0.10 & $10^{-3.5}$ \\
    100 & 140 & bitonic  & 0.10 & $10^{-3.5}$ \\
    150 & 170 & bitonic  & 0.10 & $10^{-3.5}$ \\
    200 & 200 & bitonic  & 0.10 & $10^{-4}$ \\
    \bottomrule
  \end{tabular}
\end{table}
\subsection{TSP}
\label{sec:apdx_TSP_details}
\paragraph{Dataset} 
For TSP-20 and TSP-50, we use the dataset from \citet{zhang2025symmetricdiffuserslearningdiscretediffusion}, which is adapted from \citet{joshi2021learning}. The training set contains $1{,}512{,}000$ graphs, and the test set contains $1{,}280$ graphs.

\paragraph{Model architecture}
For \SD{}, we use a 16-layer encoder-only Transformer with hidden dimension 1024, feedforward dimension 256, with 8 attention heads. For \methodname{}, we match the overall model capacity by using a 16-layer encoder--decoder Transformer, with 8 layers in the encoder and 8 layers in the decoder. We keep the hidden and feedforward dimensions the same as \SD{}.

\paragraph{Training}
All models are trained on a single NVIDIA H100 GPU for 50 epochs with batch size 64. We use a peak learning rate of $2\times 10^{-4}$ with a cosine decay schedule and $51{,}600$ warm-up steps.

\paragraph{Inference and metrics}
At inference time, we evaluate on the $1{,}280$ held-out test graphs and report two metrics: the tour length of the predicted solution, $L_\text{pred}$, and the optimality gap relative to an OR solver baseline, $L_\text{OR}$. The optimality gap is defined as
\[
\text{Gap}=\frac{L_\text{pred}-L_\text{OR}}{L_\text{OR}}
\]

\section{Derivations}
\label{sec:apdx_bridge_kernel_derivation}

Here we derive the reverse conditional transition kernel $q(z_s\mid z_t,z_0,z_1)$ for $0<s<t<1$ in Eq.~\ref{eq:bridge_backward_kernel}. We present the derivation for a single scalar soft-rank coordinate; the $N$-dimensional extension follows by independence across coordinates.

For the (unreflected) Brownian bridge with noise scale $\eta$, the marginal at time $u\in(0,1)$ is Gaussian:
\begin{equation}
q(z_u \mid z_0,z_1)
=\mathcal{N}\!\bigl(z_u;\,\mu(u),\,\eta^2 u(1-u)\bigr),
\qquad
\mu(u)\coloneqq (1-u)z_0+u z_1.
\label{eq:bridge_marginal}
\end{equation}

A convenient representation of a Brownian bridge is
\begin{equation}
z_u \;=\; \mu(u) + \eta\, b_u,\qquad u\in[0,1],
\end{equation}
where $b_u$ is a standard Brownian bridge with $b_0=b_1=0$ and covariance
$\mathrm{Cov}(b_s,b_t)=\min(s,t)-st$.
Therefore, conditional on $(z_0,z_1)$, the pair $(z_s,z_t)$ is jointly Gaussian with means
\begin{equation}
\mathbb{E}[z_s\mid z_0,z_1]=\mu(s),\qquad
\mathbb{E}[z_t\mid z_0,z_1]=\mu(t),
\end{equation}
variances
\begin{equation}
v_s \coloneqq \mathrm{Var}(z_s\mid z_0,z_1)=\eta^2 s(1-s),\qquad
v_t \coloneqq \mathrm{Var}(z_t\mid z_0,z_1)=\eta^2 t(1-t),
\end{equation}
and covariance (for $s<t$)
\begin{equation}
c_{st}\coloneqq \mathrm{Cov}(z_s,z_t\mid z_0,z_1)
=\eta^2\bigl(\min(s,t)-st\bigr)
=\eta^2 s(1-t).
\end{equation}
Hence, writing $\mu_s\coloneqq \mu(s)$, $\mu_t\coloneqq \mu(t)$ and $c\coloneqq c_{st}$,
\begin{equation}
q(z_s,z_t\mid z_0,z_1)
=\mathcal{N}\!\left(
\begin{bmatrix} z_s\\ z_t\end{bmatrix};
\begin{bmatrix} \mu_s\\ \mu_t\end{bmatrix},
\begin{bmatrix} v_s & c\\ c & v_t\end{bmatrix}
\right),
\qquad
q(z_t\mid z_0,z_1)=\mathcal{N}(z_t;\mu_t,v_t).
\label{eq:joint_and_marginal}
\end{equation}

By Bayes' rule,
\begin{equation}
q(z_s\mid z_t,z_0,z_1)=\frac{q(z_s,z_t\mid z_0,z_1)}{q(z_t\mid z_0,z_1)}.
\label{eq:cond_ratio}
\end{equation}
Since the denominator does not depend on $z_s$ (for fixed $z_t$), it suffices to view the joint density as a function of $z_s$ and complete the square. Let $x\coloneqq z_s-\mu_s$ and $y\coloneqq z_t-\mu_t$. Using the closed-form inverse for a symmetric $2\times2$ matrix, the inverse of the covariance matrix is then
\begin{equation}
\begin{bmatrix} v_s & c\\ c & v_t\end{bmatrix}^{-1}
=\frac{1}{D}
\begin{bmatrix} v_t & -c\\ -c & v_s\end{bmatrix},
\qquad
D\coloneqq v_s v_t - c^2.
\end{equation}
Therefore,
\begin{align}
\begin{bmatrix}x\\y\end{bmatrix}^{\!\top}
\begin{bmatrix} v_s & c\\ c & v_t\end{bmatrix}^{-1}
\begin{bmatrix}x\\y\end{bmatrix}
&=\frac{1}{D}\left(v_t x^2 - 2cxy + v_s y^2\right)\nonumber\\
&=\frac{v_t}{D}\left(x-\frac{c}{v_t}y\right)^2 + \text{const}(y),
\end{align}
where $\text{const}(y)$ does not depend on $x$ (hence not on $z_s$). This yields the Gaussian conditional
\begin{equation}
z_s\mid z_t,z_0,z_1 \sim 
\mathcal{N}\!\left(
\mu_s+\frac{c}{v_t}(z_t-\mu_t),\;
v_s-\frac{c^2}{v_t}
\right).
\label{eq:cond_general}
\end{equation}
Substituting $v_s=\eta^2 s(1-s)$, $v_t=\eta^2 t(1-t)$, and $c=\eta^2 s(1-t)$ (for $s<t$), we obtain
\begin{equation}
\frac{c}{v_t}=\frac{s}{t},
\qquad
v_s-\frac{c^2}{v_t}=\eta^2\frac{s(t-s)}{t}.
\end{equation}
Hence,
\begin{equation}
z_s\mid z_t,z_0,z_1 \sim 
\mathcal{N}\!\left(
\mu(s)+\frac{s}{t}\bigl(z_t-\mu(t)\bigr),\;
\eta^2\frac{s(t-s)}{t}
\right),
\qquad 0<s<t<1.
\end{equation}

\paragraph{Extension to $N$ dimensions.}
Under coordinate-wise independent bridges sharing the same noise scale $\eta$, the same conditional holds with $\epsilon\sim\mathcal{N}(0,I_N)$, i.e., the conditional covariance becomes $\eta^2\frac{s(t-s)}{t}I_N$.

\section{Additional Analysis}
\label{sec:apdx_additional_analysis}

\subsection{Sensitivity to the initial reference distribution}
\label{sec:apdx_init_dist}

A natural design question for \methodname{} is the choice of the reference distribution $p_\text{ref}$ used as the endpoint of the forward Brownian bridge (Eq.~\ref{eq:forward_bridge}). Two simple choices are the unit-cube uniform $\mathrm{Unif}([0,1]^N)$ (matching the natural domain of soft ranks) and the standard Gaussian $\mathcal{N}(0, I_N)$ (matching the convention from continuous-space diffusion models in $\mathbb{R}^N$). We ablate this choice on CIFAR-10 jigsaw with $4\times 4$ ($N=16$) for both the cGPL and Pointer-cGPL decoders.

\paragraph{Matched train and evaluation reference.}
We first compare the two reference choices in the \emph{matched} setting, where training and evaluation use the same $p_\text{ref}$ (Table~\ref{tab:init-matched}). Performance is essentially identical across uniform and Gaussian choices for both decoders (within $0.5$ percentage points of exact-match accuracy and $0.002$ in Kendall $\tau$), indicating that under matched train/evaluation, the choice of $p_\text{ref}$ is not a critical hyperparameter.

\begin{table}[!htb]
\centering
\small
\caption{Matched train/eval $p_\text{ref}$ on CIFAR10 jigsaw ($n=4$, $\sigma=0.5$, $T=5$). Both Uniform and Gaussian reference distributions yield essentially identical performance, indicating the choice of $p_\text{ref}$ has negligible impact on training.}
\label{tab:init-matched}
\begin{tabular}{@{}llccc@{}}
\toprule
Decoder & Init Distribution & Perm Acc (\%) & Kendall $\tau$ & Prop Correct (\%) \\
\midrule
\multirow{2}{*}{cGPL}        & Uniform  & 86.67 & 0.9303 & 93.93 \\
                             & Gaussian & 86.96 & 0.9317 & 94.08 \\
\midrule
\multirow{2}{*}{Pointer-cGPL} & Uniform  & 87.35 & 0.9365 & 94.43 \\
                              & Gaussian & 86.94 & 0.9347 & 94.38 \\
\bottomrule
\end{tabular}
\end{table}

\paragraph{Mismatched train and evaluation reference.}
A stronger test of robustness is to deliberately \emph{mismatch} the training and evaluation reference distributions—training under one distribution and evaluating under the other (Table~\ref{tab:init-mismatched}). All four train/eval combinations yield essentially the same accuracy (within noise).

\begin{table}[!htb]
\centering
\small
\caption{Mismatched train/eval $p_\text{ref}$ on CIFAR10 jigsaw ($n=4$, $\sigma=0.5$, $T=5$). Each cell shows Perm Acc / Prop Correct / Kendall $\tau$. Performance is preserved even when training and evaluation use different reference distributions, confirming the Brownian bridge formulation drives $z$ toward $z_0$ regardless of starting distribution.}
\label{tab:init-mismatched}
\begin{tabular}{@{}llcc@{}}
\toprule
Decoder & Train $p_\text{ref}$ & Eval: Uniform & Eval: Gaussian \\
\midrule
\multirow{2}{*}{cGPL}         & Uniform  & 86.67 / 93.93 / $\tau\!=\!0.930$ & 87.09 / 94.08 / $\tau\!=\!0.931$ \\
                              & Gaussian & 86.71 / 94.01 / $\tau\!=\!0.932$ & 86.96 / 94.08 / $\tau\!=\!0.932$ \\
\midrule
\multirow{2}{*}{Pointer-cGPL} & Uniform  & 87.35 / 94.43 / $\tau\!=\!0.937$ & 87.33 / 94.51 / $\tau\!=\!0.937$ \\
                              & Gaussian & 87.17 / 94.31 / $\tau\!=\!0.935$ & 86.94 / 94.38 / $\tau\!=\!0.935$ \\
\bottomrule
\end{tabular}
\end{table}

We adopt $\mathrm{Unif}([0,1]^N)$ in the main body of the paper for simplicity (matching the natural soft-rank domain), but the results above show that this choice is not critical to the performance of \methodname{}, and practitioners can swap in $\mathcal{N}(0,I_N)$ or other reasonable references without affecting performance.

\subsection{Discussion on alternative choices of relaxation space}
\label{sec:apdx_alternative_spaces}

A natural perspective on permutation generation is to relax \emph{permutation matrices} $P_\sigma \in \{0,1\}^{N\times N}$ rather than ranks. The two most common matrix-based relaxations are the \textbf{Birkhoff polytope} $B_N$ (the convex hull of permutation matrices, i.e., doubly stochastic matrices)~\citep{mena2018learninglatentpermutationsgumbelsinkhorn,cuturi2019differentiablerankssortingusing} and the \textbf{orthogonal group} $O(N)$. We chose instead to relax the rank vector to $[0,1]^N$ for both representational and computational reasons. Table~\ref{tab:relaxation-compare} contrasts the three spaces.

\begin{table}[h]
\centering
\small
\caption{Comparison of three continuous relaxations of permutations. The soft-rank relaxation $[0,1]^N$ used by \methodname{} has linear dimension and admits independent closed-form Brownian-bridge dynamics, while Birkhoff- and $O(N)$-based relaxations require coupled, more expensive constructions.}
\label{tab:relaxation-compare}
\begin{tabular}{@{}p{2.8cm}p{4.0cm}p{4.0cm}p{4.0cm}@{}}
\toprule
& $[0,1]^N$ (ours) & Birkhoff polytope $B_N$ & Orthogonal group $O(N)$ \\
\midrule
Dimension & $N$ & $(N-1)^2$ & $N(N-1)/2$ \\
\midrule
Dynamics & Independent reflected Brownian bridges with closed-form transition kernels & Coupled entries; requires per-step Sinkhorn projection & Riemannian geodesics \\
\midrule
Decoding & $\mathrm{argsort}$ (Eq.~\ref{eq:map_to_grid_def}) & Hungarian or Sinkhorn rounding & Nearest permutation matrix \\
\bottomrule
\end{tabular}
\end{table}

One key practical advantage is the dimensionality scaling. Our representation requires only $N$ scalars, whereas $B_N$ requires $(N-1)^2$ free parameters; at $N=200$ this is a factor of $\sim$200$\times$ ($200$ vs.\ $39{,}601$ dimensions). This quadratic scaling makes matrix-based relaxations increasingly impractical for long sequences, both for the computational cost of the diffusion dynamics and for the modeling burden of the score / denoising network. Furthermore, the entries of a doubly stochastic matrix are constrained by row- and column-sum identities, so a diffusion process in $B_N$ must either project back onto $B_N$ at each step (e.g., via Sinkhorn iteration~\citep{sinkhorn1964relationship,mena2018learninglatentpermutationsgumbelsinkhorn}) or operate in a tangent space; both incur per-step cost that is absent from our coordinate-wise reflected bridge.

We emphasize that this is a difference of \emph{representation}, not a competing relaxation: rather than softening permutation matrices, we soften the rank vector itself, an equivalent encoding of the permutation that scales linearly in $N$ and admits independent dynamics.

\paragraph{Extension to constrained permutations.}
A further benefit of working in $[0,1]^N$ is that structural constraints on permutations translate into simple \emph{geometric} constraints on the latent space: common combinatorial constraints (partial orders, forbidden assignments) reduce to half-space inequalities or coordinate masks. For example, the partial-order constraint ``$a$ appears before $b$'' becomes the half-space $\{z\in[0,1]^N : z_a < z_b\}$, which our reflected bridge preserves by adding a single reflection plane, whereas the same constraint couples all $N^2$ entries in the Birkhoff polytope. We do not pursue constrained permutation generation experimentally and leave this direction to future work.

\subsection{Chamber geometry of the soft-rank latent space}
\label{sec:apdx_chamber}

The decoding map $\sigma = \mathrm{argsort}(\mathrm{argsort}(Z))$ used by \methodname{} is \emph{many-to-one}: many distinct latent vectors $Z\in[0,1]^N$ induce the same permutation $\sigma$. This subsection elaborates on the geometry of this map and argues that the many-to-one structure is a useful feature of our parameterization rather than a limitation.

\paragraph{Chambers.}
The latent cube $[0,1]^N$ is partitioned (up to a measure-zero boundary where coordinates tie) into $N!$ open convex \emph{chambers}, one per permutation:
\begin{equation}
C_\sigma \;\coloneqq\; \big\{z \in [0,1]^N \,:\, z_{\sigma^{-1}(1)} < z_{\sigma^{-1}(2)} < \cdots < z_{\sigma^{-1}(N)}\big\}.
\end{equation}
The chambers $\{C_\sigma\}_{\sigma\in\mathcal{S}_N}$ are congruent (related by coordinate permutations) and partition the open cube; each has volume $1/N!$, so the uniform reference distribution $\mathrm{Unif}([0,1]^N)$ assigns equal mass to every permutation—a natural uninformative prior on $\mathcal{S}_N$. The $\mathrm{LiftToGrid}$ operator (Eq.~\ref{eq:map_to_grid_def}) places the data latent $Z_0$ at the centroid of $C_{\sigma_0}$, the point of maximum distance ($1/(2N)$) to every chamber boundary, providing the most boundary-robust starting point for forward noising.

\paragraph{Many-to-one decoding.}
Three consequences of the chamber structure make $[0,1]^N$ a more forgiving latent space than zero-dimensional vertex-based representations such as $\{P_\sigma\} \subset B_N$:

\begin{enumerate}
\item \textbf{Tolerance margin.} The reverse-time sampler only needs to place its prediction $\hat z_0$ \emph{anywhere inside} the correct chamber $C_{\sigma_0}$—a positive-volume region. Contrast this with Birkhoff-based methods, where the targets $P_\sigma$ are zero-dimensional vertices of $B_N$ and correct recovery relies on a separate rounding procedure (e.g., the Hungarian algorithm~\citep{kuhn1955hungarian}) to map nearby doubly stochastic matrices to the correct vertex. The chamber margin grants \methodname{} robustness to small errors in the denoising prediction.
\item \textbf{Confidence refinement.} Within a chamber, the reverse trajectory may move \emph{toward the interior} (away from boundaries) without changing the decoded permutation. This provides a natural form of confidence calibration: after the chamber is selected, further reverse steps reduce boundary distance, increasing margin without altering the answer.
\item \textbf{Tie-breaking is measure-zero.} The score network outputs continuous values, so $\Pr(z_i = z_j) = 0$ almost surely under any continuous noise model. Decoding ambiguities at chamber boundaries therefore occur with probability zero, and the empirical predictions of our trained networks consistently maintain clear separation from boundaries.
\end{enumerate}

Together, these properties suggest that the apparent ``redundancy'' of having $N!$ chambers map to the same set of $N!$ permutations is not a wasteful overparameterization, but a deliberate trade of zero-dimensional precision for positive-volume tolerance that the diffusion sampler can exploit.

\section{Additional Figures and Tables}
\label{sec:apdx_additional_figures_and_tables}
\begin{figure}[h]
  \centering
  \begin{subfigure}[c]{0.4\linewidth}
    \centering
    \vspace{0pt}
    \includegraphics[width=\linewidth]{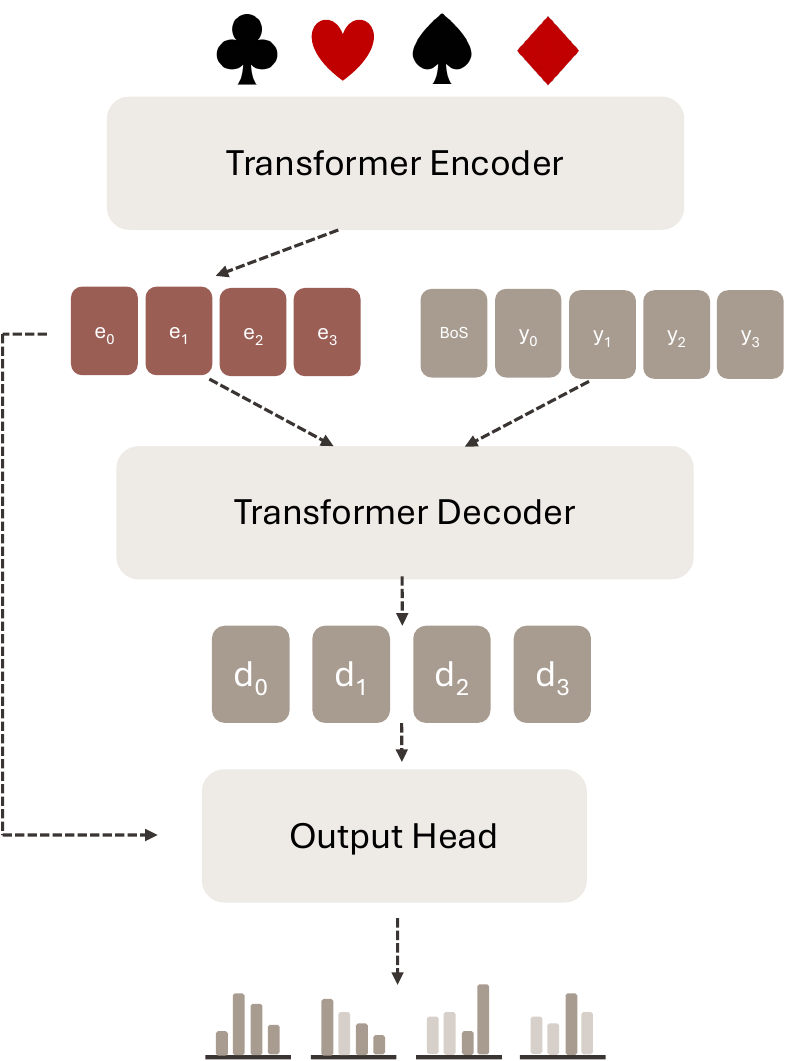}
    \caption{Model Architecture.}
    \label{fig:subfig_model_arch}
  \end{subfigure}%
  \hfill
  \begin{subfigure}[c]{0.3\linewidth}
    \centering
    \vspace{0pt}
    \includegraphics[width=\linewidth]{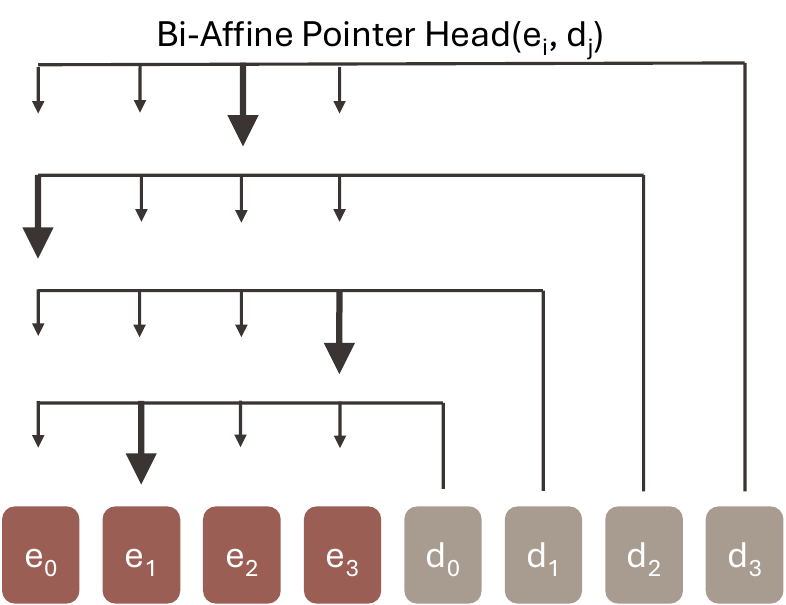}
    \caption{Pointer-cGPL factorization.}
    \label{fig:subfig_ptr}
  \end{subfigure}
  \caption{\textbf{Model architecture and Pointer-cGPL parameterization for permutation generation.}We adopt a standard encoder--decoder Transformer backbone \cite{vaswani2023attentionneed}. In the vanilla cGPL parameterization, the decoder states are mapped to logits via a linear output head (yielding a distribution over candidates at each step). In contrast, Panel~\ref{fig:subfig_ptr} illustrates \emph{Pointer-cGPL}, where a bi-affine compatibility module scores each encoded item representation against the current decoder state, producing step-wise logits over the input items that can be interpreted as a pointer distribution.}
  \label{fig:model_arch}
\end{figure}

\begin{table*}[!htb]
\centering
\small
\caption{Ablation study over forward diffusion processes, reverse models, and parametrizations on sorting 32 4-digit MNIST images.}
\makebox[\textwidth][c]{%
\begin{tabular}{@{}llllccc@{}}
\toprule
Forward Process & Reverse Model & Parametrization
& \KT{} $\uparrow$ & Accuracy $\uparrow$ & Correctness $\uparrow$ \\
\midrule
\multirow{2}{*}{Riffle Shuffle}
& \multirow{2}{*}{GPL}
& $x_{t-1}$ & 0.8610 & 0.5896 & 0.8846 \\
& & $x_{0}$   & -0.0013 & 0.0000 & 0.0312 \\
\midrule
\multirow{2}{*}{\methodname{}}
& cGPL
& $x_{t-1}$ & 0.8764 & 0.6425 & 0.8958 \\
& cGPL w/ Biaffine Pointer
& $x_{t-1}$ & 0.9074 & 0.7152 & 0.9232 \\
\midrule
\multirow{4}{*}{Riffle Shuffle}
& cGPL
& $x_{t-1}$ & 0.8527 & 0.5833 & 0.8758 \\
& cGPL w/ Biaffine Pointer
& $x_{t-1}$ & 0.8742 & 0.6404 & 0.8950 \\
& cGPL
& $x_{0}$   & -0.0011 & 0.0000 & 0.0313 \\
& cGPL w/ Biaffine Pointer
& $x_{0}$   & 0.0001 & 0.0000 & 0.0314 \\
\bottomrule
\end{tabular}%
}
\label{tab:ablation-table}
\end{table*}

\begin{figure}[!htb]
    \centering
    \includegraphics[width=0.8\linewidth]{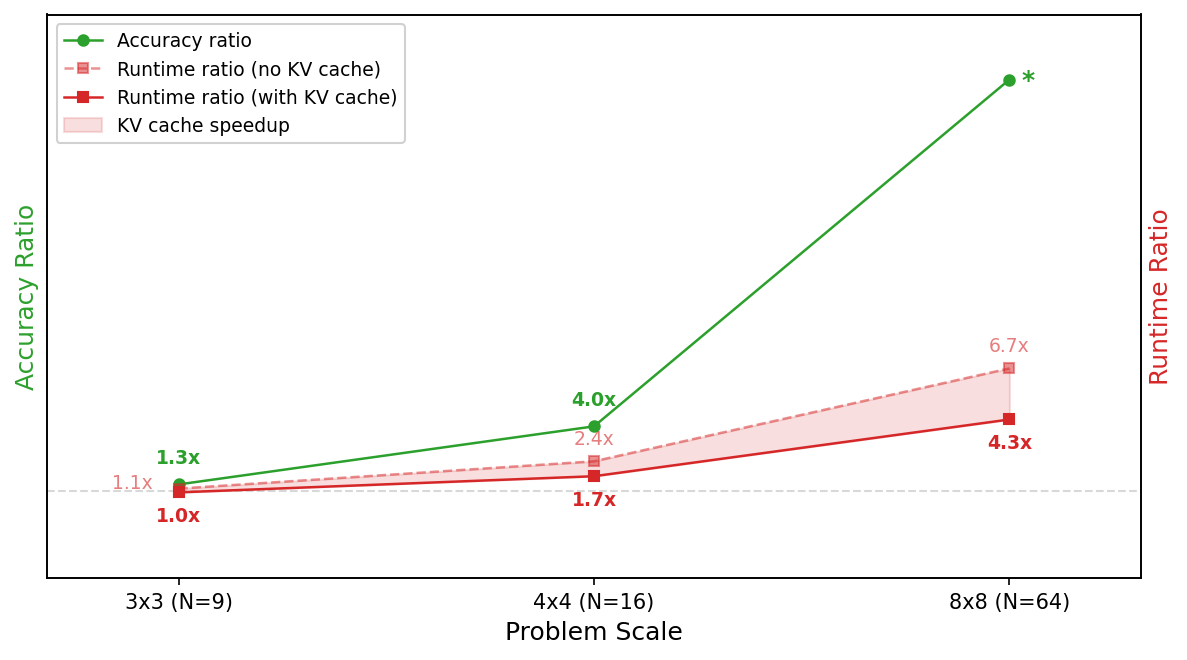}
    \caption{\textbf{Accuracy ratio} (green; Soft-Rank Pointer accuracy / SymmetricDiffusers accuracy) and \textbf{runtime ratio} (red; Soft-Rank Pointer wall-clock time / SymmetricDiffusers wall-clock time) across problem scales. The accuracy advantage grows superlinearly with $N$, reaching $4.0\times$ at $N\!=\!16$ and becoming unbounded at $N\!=\!64$ (*baseline accuracy is 0\%). Runtime overhead remains moderate: with KV cache optimization (solid red), the slowdown is $1.0\times$--$4.3\times$, while the shaded region shows the speedup across diffusion steps (up to $1.55\times$ at $N\!=\!64$). \textbf{The accuracy gains far outpace the runtime cost at every scale.}}
      \label{fig:accuracy_runtime_kvcache}
  \end{figure}
\begin{figure}[t]
    \centering
    \includegraphics[width=0.8\linewidth]{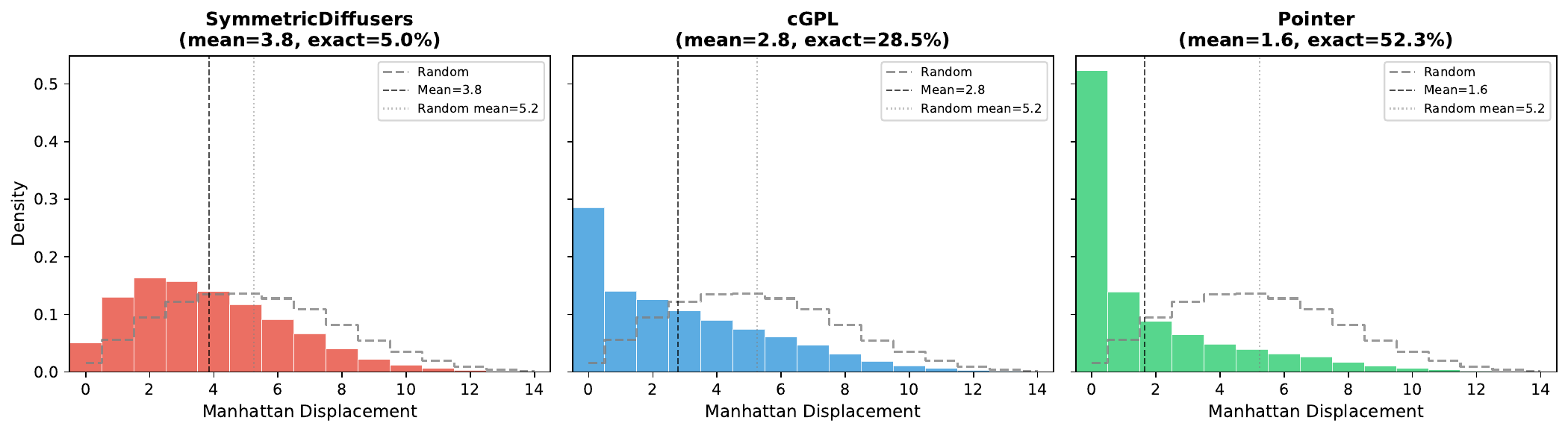}
    \caption{\textbf{Patch-wise Manhattan displacement distributions on $8\times8$ CIFAR-10 jigsaw.} Manhattan displacement is the grid distance between a patch's predicted and correct positions. Dashed gray = random baseline (mean $5.2$). Left: \SD{} (mean $3.8$). Middle: cGPL (mean $2.8$). Right: Pointer-cGPL (mean $1.6$).} 
    \label{fig:displacement_histogram}
\end{figure}
\begin{table}[!htb]
\centering
\small
\caption{\textbf{Wall-clock inference time.} Single H100 GPU, $10{,}000$ test samples.}
\label{tab:runtime}
\begin{tabular}{@{}lccc@{}}
\toprule
Method & $3\!\times\!3$ & $4\!\times\!4$ & $8\!\times\!8$ \\
\midrule
\SD{} & 28.0s & 21.6s & 36.3s \\
\midrule
cGPL, w/o cache  & 31.4s & 41.5s & 216.6s \\
cGPL, w/ cache   & 26.3s & 31.4s & 131.7s \\
\hspace{1em}\textit{speedup} & \textit{1.19}$\times$ & \textit{1.32}$\times$ & \textit{1.64}$\times$ \\
\midrule
Pointer-cGPL, w/o cache & 31.6s & 51.6s & 242.3s \\
Pointer-cGPL, w/ cache  & 26.7s & 36.7s & 156.7s \\
\hspace{1em}\textit{speedup} & \textit{1.18}$\times$ & \textit{1.41}$\times$ & \textit{1.55}$\times$ \\
\bottomrule
\end{tabular}
\end{table}

\end{document}